\documentclass[runningheads,a4paper]{llncs}

\usepackage[utf8]{inputenc}
\usepackage{amsmath,amssymb}
\usepackage{paralist}
\usepackage{color}
\usepackage[detect-weight=true, binary-units=true]{siunitx}
\usepackage{url}

\usepackage{calc}
\usepackage{amssymb}
\usepackage{multirow}
\usepackage{microtype} 
\usepackage{colortbl,etoolbox,xcolor} 
\robustify\bfseries 

\usepackage{booktabs}
\usepackage{alltt}
\usepackage[caption=false]{subfig}

\usepackage[normalem]{ulem} 

\usepackage{algorithm,algpseudocode}
\MakeRobust{\Call}
\algrenewcommand\algorithmicindent{1.0em}%

\usepackage{pgfplots}
\usepackage{pgfplotstable}
\usepgfplotslibrary{statistics}
\pgfplotsset{compat=1.13}
\pgfplotsset{
    dne/.style 2 args={
        x filter/.append code={
            \edef\tempa{\thisrow{#1}}
            \edef\tempb{#2}
            \ifx\tempa\tempb
            \else
                
            \fi
        },
    }
}
\usepgfplotslibrary{groupplots}
\pgfplotscreateplotcyclelist{collist}{
    {color=red},
    {color=blue},
    {color=green},
    {color=black},
    {color=brown},
    {color=purple}
}

\usepackage{needspace}

\begin{document}

\mainmatter

\title{Observing the Population Dynamics in GE\\by means of the Intrinsic Dimension}
\titlerunning{Observing the Population Dynamics in GE with ID}

\author{
    Eric Medvet\inst{1} \and
    Alberto Bartoli\inst{1} \and
    Alessio Ansuini\inst{2} \and
    Fabiano Tarlao\inst{1}}
\institute{
    Department of Engineering and Architecture, University of Trieste, Trieste, Italy \and
    International School for Advanced Studies (SISSA/ISAS), Trieste 34136, Italy
}

\maketitle

\begin{abstract}
    We explore the use of Intrinsic Dimension (ID) for gaining insights in how populations evolve in Evolutionary Algorithms.
    ID measures the minimum number of dimensions needed to accurately describe a dataset and its estimators are being used more and more in Machine Learning to cope with large datasets.
    We postulate that ID can provide information about population which is complimentary w.r.t.\ what (a simple measure of) diversity tells.
    We experimented with the application of ID to populations evolved with a recent variant of Grammatical Evolution.
    The preliminary results suggest that diversity and ID constitute two different points of view on the population dynamics.
\end{abstract}


\section{Introduction and related work}
\label{sec:introduction}
In Evolutionary Computation (EC) a population of candidate solutions to a problem, also known as \emph{individuals}, evolves by mutating and recombining fitter individuals.
It is well known among EC scholars and practitioners that if the population is composed of many very similar individuals, i.e., if it is not \emph{diverse}, the exploration phase of the evolution may be ineffective, eventually leading to the so called \emph{premature convergence}.
Lack of diversity has been acknowledged as a problem of Evolutionary Algorithms (EAs) since the beginning of the EC research field and has been widely studied~\cite{squillero2016divergence}.
In particular, different ways have been proposed for measuring the population \emph{diversity}.

In this work, we explore the use of the Intrinsic Dimension (ID) for characterizing the structure of the population during an EA run and understanding if and how it relates with diversity.
ID can be seen as the minimum number of variables which are needed to accurately describe the features of a system~\cite{facco2017estimating} and has been used in different domains in which data is, in general, large.
Often, ID has been proposed as a mean for reducing the dimensionality of the data with the aim of fighting the so called ``curse of dimensionality''~\cite{campadelli2015intrinsic}.

In the EC context, we postulate that measuring the ID of the population may help the understanding of the structure of the population in a more detailed way than measuring its diversity.
The intuition is that the action of the selective pressure may impose a ``shape'' to the population during the evolution and that that shape corresponds to a region of the search space of which dimension is much smaller than the full space dimension.
From another point of view, during the evolution the individuals may move along one or more paths within the \emph{fitness landscape}.

In order to gain insights about our speculation, we performed a set of experiments in which we considered a popular EA, Grammatical Evolution (GE)~\cite{ryan1998grammatical}, and measured how the ID, the diversity, and the fitness of the best individual vary over the evolution.
We considered GE for three reasons:
\begin{inparaenum}[(a)]
    \item its suitability for any problem which can be described by a Context Free Grammar makes it an EA of great practical relevance;
    \item it employs a bit string-based representation, hence the search space allows using a simple distance among genotypes (i.e., the Hamming distance between bit strings);
    \item it has been showed to suffer from lack of diversity~\cite{medvet2017comparative} and some methods for preserving it has been proposed too~\cite{medvet2017effective}.
\end{inparaenum}

From our preliminary experiments, we found that ID seems to provide a complementary information w.r.t.\ diversity, the latter measured as the fraction of unique genotypes in the population.
In particular, ID always increases during the evolution, whereas the diversity initially quickly decreases and later slowly increases when the problem remains unsolved.
This finding suggests that the high diversity in the initial population might be of ``poor quality'', since it does not correspond to an high ID---a finding which confirms the relevance of the population initialization in GE~\cite{nicolau2017understanding}.

As the lack of diversity is perceived as a key problem with practical effects in the EC community, many studies have been devoted to it.
In particular, different ways for measuring the diversity have been proposed---for space constraints, we refer the reader to~\cite{squillero2016divergence} for a comprehensive analysis.
To the best of our knowledge, no previous works exist on the application of the ID to a population of an EA.

\section{Background: estimated Intrinsic Dimension}
\label{sec:intrinsic-dimension}
The ID of a dataset, i.e., a set of points defined in a high-dimensional space, can be seen as the minimum number of variables which are needed to accurately describe that dataset.
Measuring the actual ID of a real dataset may be an hard task, due to the presence of noise.

In order to deal with noise and to reduce the computational burden of computing the actual ID, many ways of estimating it have been proposed.
We here use TWO-NN, a recent estimation method which requires to compute only the distance of each point to its two nearest neighbors (NNs)~\cite{facco2017estimating}.

In detail, TWO-NN works as follows.
Let $P$ be the set of $n$ items whose ID has to be computed and let $D$ be the corresponding distance matrix, i.e., $d_{i,j}$ is the distance between the $i$th item $p_i \in P$ and the $j$th item $p_j \in P$.
First, the ratios $\vec{r}=(r_1, \dots, r_n)$ between the distances of the two closest items for each item is computed as $r_i = \min_{j \ne i,h \ne i} \frac{d_h}{d_j}$, with $d_h \ge d_j$.
Second, $\vec{r}$ is sorted in ascending order.
Third, the values of $\vec{x} = \log \vec{r}$ and $\vec{y} = -\log(1-\frac{1}{n} \{0, 1, \dots, n-1\})$ are computed.
Fourth and finally, the points defined by $\vec{x},\vec{y}$ are fitted with a straight line and the resulting slope is taken as the ID of the points in $P$.

In our work, $P$ is composed by a population of bit strings observed during an EA run.
This exacerbates the estimation of the ID for two reasons:
\begin{inparaenum}[(i)]
    \item the Hamming distance among individuals is discrete, rather than continuous;
    \item $P$ may contain many duplicated individuals, due to low diversity in later stages of the evolution.
\end{inparaenum}
As a result, the quality of the linear fitting of the $\vec{x},\vec{y}$ points, expressed in terms of the product-moment coefficient of correlation, can be low.
In order to mitigate the latter cause, we computed the ID on the set of unique genotypes, rather than on the entire population.

\section{Experiments, results, and discussion}
We experimented with a recent variant of GE, called Weighted Hierarchical GE (WHGE)~\cite{medvet2017hierarchical} which differs from the original only in the representation internals and has been shown to perform better.
We considered two benchmark problems with tunable hardness: the Parity problem, for which we experimented with the number of input bits in $b \in \{3, \dots, 7\}$, and the KLandscape synthetic benchmark~\cite{vanneschi2011k}, for which we set $k \in \{3, \dots, 7\}$.
For both, the lower the parameter, the easier the problem.

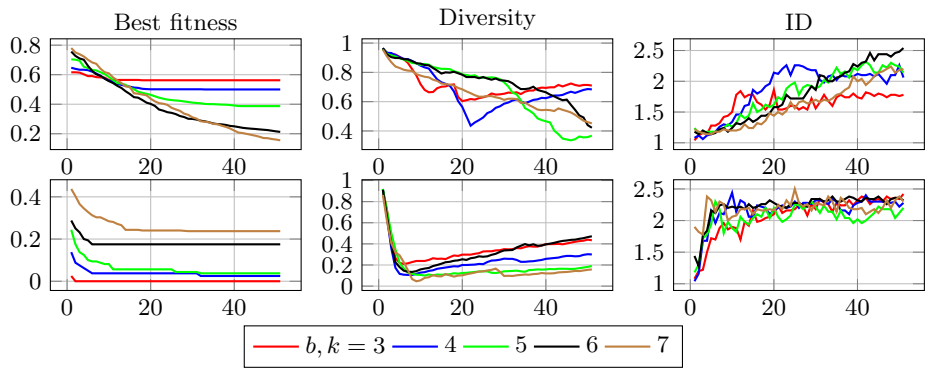
\begin{figure}
    \centering
    \begin{tikzpicture}
        \begin{groupplot}[
            width=.40\linewidth,
            height=.25\linewidth,            
            no markers,
            grid=major,
            title style={yshift=-1.5ex},
            every axis plot/.append style={thick},
            cycle list name=collist,
            group style={
                group size=3 by 2,
                horizontal sep=8mm,
                vertical sep=4mm
            }
        ]
            \nextgroupplot[title=Best fitness]
            \addplot table[x=iterations,y=bf,dne={k}{3}]{data/kland.txt};
            \addplot table[x=iterations,y=bf,dne={k}{4}]{data/kland.txt};
            \addplot table[x=iterations,y=bf,dne={k}{5}]{data/kland.txt};
            \addplot table[x=iterations,y=bf,dne={k}{6}]{data/kland.txt};
            \addplot table[x=iterations,y=bf,dne={k}{7}]{data/kland.txt};
            \nextgroupplot[title=Diversity]
            \addplot table[x=iterations,y=d,dne={k}{3}]{data/kland.txt};
            \addplot table[x=iterations,y=d,dne={k}{4}]{data/kland.txt};
            \addplot table[x=iterations,y=d,dne={k}{5}]{data/kland.txt};
            \addplot table[x=iterations,y=d,dne={k}{6}]{data/kland.txt};
            \addplot table[x=iterations,y=d,dne={k}{7}]{data/kland.txt};
            \nextgroupplot[title=ID,
                legend columns=5,
                legend to name=legend,
                legend entries={{$b,k=3$},4,5,6,7}]
            \addplot table[x=iterations,y=id,dne={k}{3}]{data/kland.txt};
            \addplot table[x=iterations,y=id,dne={k}{4}]{data/kland.txt};
            \addplot table[x=iterations,y=id,dne={k}{5}]{data/kland.txt};
            \addplot table[x=iterations,y=id,dne={k}{6}]{data/kland.txt};
            \addplot table[x=iterations,y=id,dne={k}{7}]{data/kland.txt};
            \nextgroupplot
            \addplot table[x=iterations,y=bf,dne={k}{3}]{data/parity.txt};
            \addplot table[x=iterations,y=bf,dne={k}{4}]{data/parity.txt};
            \addplot table[x=iterations,y=bf,dne={k}{5}]{data/parity.txt};
            \addplot table[x=iterations,y=bf,dne={k}{6}]{data/parity.txt};
            \addplot table[x=iterations,y=bf,dne={k}{7}]{data/parity.txt};
            \nextgroupplot
            \addplot table[x=iterations,y=d,dne={k}{3}]{data/parity.txt};
            \addplot table[x=iterations,y=d,dne={k}{4}]{data/parity.txt};
            \addplot table[x=iterations,y=d,dne={k}{5}]{data/parity.txt};
            \addplot table[x=iterations,y=d,dne={k}{6}]{data/parity.txt};
            \addplot table[x=iterations,y=d,dne={k}{7}]{data/parity.txt};
            \nextgroupplot
            \addplot table[x=iterations,y=id,dne={k}{3}]{data/parity.txt};
            \addplot table[x=iterations,y=id,dne={k}{4}]{data/parity.txt};
            \addplot table[x=iterations,y=id,dne={k}{5}]{data/parity.txt};
            \addplot table[x=iterations,y=id,dne={k}{6}]{data/parity.txt};
            \addplot table[x=iterations,y=id,dne={k}{7}]{data/parity.txt};
        \end{groupplot}
    \end{tikzpicture}
    \\
    \ref{legend}
    \caption{
        Results, averaged across the $5$ runs, for Klandscapes (above) and Parity (below).
    }
    \label{fig:plots}
\end{figure}

For each problem, we executed $5$ independent runs and measured, at each generation, the fitness of the best individual, the population diversity (i.e., the fraction of unique genotypes), and the population ID.
We used the standard $m+n$ replacement strategy with $m=n=500$ individuals, which we evolved for $50$ generations, and a tournament selection with size $3$.
We set the genotype size to $\SI{256}{\bit}$.
Figure~\ref{fig:plots} shows the results for the two problems, Klandscapes (above) and Parity (below)---for the best fitness, the lower, the better.

It can be seen the way ID and diversity vary during the evolution is different.
The diversity starts at $\approx 1$ in the initial population and then decreases: for Parity, the decrease is very rapid, whereas it is slower for Klandscapes; in both cases, the diversity reaches a minimum and inverts the trend when the best fitness ceases to (significantly) improve.
Instead, ID starts at $\approx 1$ and steadily increases: when diversity decreases fast, ID increases rapidly; otherwise it increases slowly.

We believe that the diversity starts increasing when the evolution stagnates, i.e., when the best fitness stops improving, because of the representation degeneracy (the tendency of mapping different genotypes to the same phenotype~\cite{medvet2017comparative}): many different genotypes mapping to the same best phenotype start to fill the population.
Interestingly, ID seems not to be affected by this phenomenon, as it remains roughly steady in later stages of the stagnating evolutions.

Moreover, final values of ID are always greater than initial values.
No sharp conclusions can be drawn in these respects, but we believe that ID values suggest that the final population tend to lie on flat regions in the search space.
On the other hand, the fact that the highly diverse initial populations exhibit a low ID suggests that the population initialization is indeed a critical component of GE, as already observed~\cite{nicolau2017understanding}.

\bibliographystyle{splncs}
\bibliography{bibliography}

\end{document}